\documentclass[conference]{IEEEtran}
\IEEEoverridecommandlockouts
\usepackage{orcidlink}
\usepackage{cite}
\usepackage{lipsum}
\usepackage{amsmath,amssymb,amsfonts}
\usepackage{algorithmic}
\usepackage{graphicx}
\usepackage{textcomp}
\usepackage{xcolor}
\usepackage{colortbl}
\usepackage{soul}

\usepackage{tikz}
\usetikzlibrary{calc}
\usetikzlibrary{spy}
\usepackage{hyperref}
\hypersetup{
    colorlinks=true,
    citecolor=blue,
    linkcolor=purple,
    filecolor=magenta,      
    urlcolor=blue,
    unicode=true,
    }
\usepackage{ragged2e} 
\usepackage{subcaption} 
\def\BibTeX{{\rm B\kern-.05em{\sc i\kern-.025em b}\kern-.08em
    T\kern-.1667em\lower.7ex\hbox{E}\kern-.125emX}}

\definecolor{mygreen}{rgb}{0.88, 1.0, 0.84} 
\definecolor{myyellow}{rgb}{1.0, .975, 0.82} 
\newcommand{\li}{{LiDAR }}
\newcommand{\includegraphicsright}[2][]{%
    \sbox0{\includegraphics{#2}}
    \includegraphics[#1, clip, viewport={0.5\wd0} 0 {\wd0} {\ht0}]{#2}
}

\begin{document}

\title{Physically Based Neural LiDAR Resimulation \\

\thanks{%
    \copyright~2025 IEEE. Personal use of this material is permitted. Permission from IEEE must be obtained for all other uses, in any current or future media, including reprinting/republishing this material for advertising or promotional purposes, creating new collective works, for resale or redistribution to servers or lists, or reuse of any copyrighted component of this work in other works.
  }%

\thanks{Richard Marcus was supported by the  (Bavarian Research Foundation) AZ-1423-20.
The authors gratefully acknowledge the scientific support and HPC resources provided by the Erlangen National High Performance Computing Center (NHR@FAU) of the Friedrich-Alexander-Universität Erlangen-Nürnberg (FAU) under the NHR project b204dc. NHR funding is provided by federal and Bavarian state authorities. NHR@FAU hardware is partially funded by the German Research Foundation (DFG) – 440719683.
}
}
\author{
\IEEEauthorblockN{Richard Marcus\orcidlink{0000-0002-6601-6457} 
and Marc Stamminger\orcidlink{0000-0001-8699-3442}}
\IEEEauthorblockA{Visual Computing Erlangen, Friedrich-Alexander-Universität Erlangen-Nürnberg, Erlangen, Germany\\
}
}

\maketitle

\begin{abstract}
Methods for Novel View Synthesis (NVS) have recently found traction in the field of \li simulation and large-scale 3D scene reconstruction.
While solutions for faster rendering or handling dynamic scenes have been proposed,  \li specific effects remain insufficiently addressed.

By explicitly modeling sensor characteristics such as rolling shutter, laser power variations, and intensity falloff, our method achieves more accurate \li simulation compared to existing techniques. We demonstrate the effectiveness of our approach through quantitative and qualitative comparisons with state-of-the-art methods, as well as ablation studies that highlight the importance of each sensor model component. 

Beyond that, we show that our approach exhibits advanced resimulation capabilities, such as generating high resolution \li scans in the camera perspective.
Our code and the resulting dataset are available at \url{https://github.com/richardmarcus/PBNLiDAR}.
\end{abstract}


\section{Introduction}
\begin{figure}[tbp]
    \centering
    \includegraphics[width=\linewidth, trim=0 50 0 38, clip]{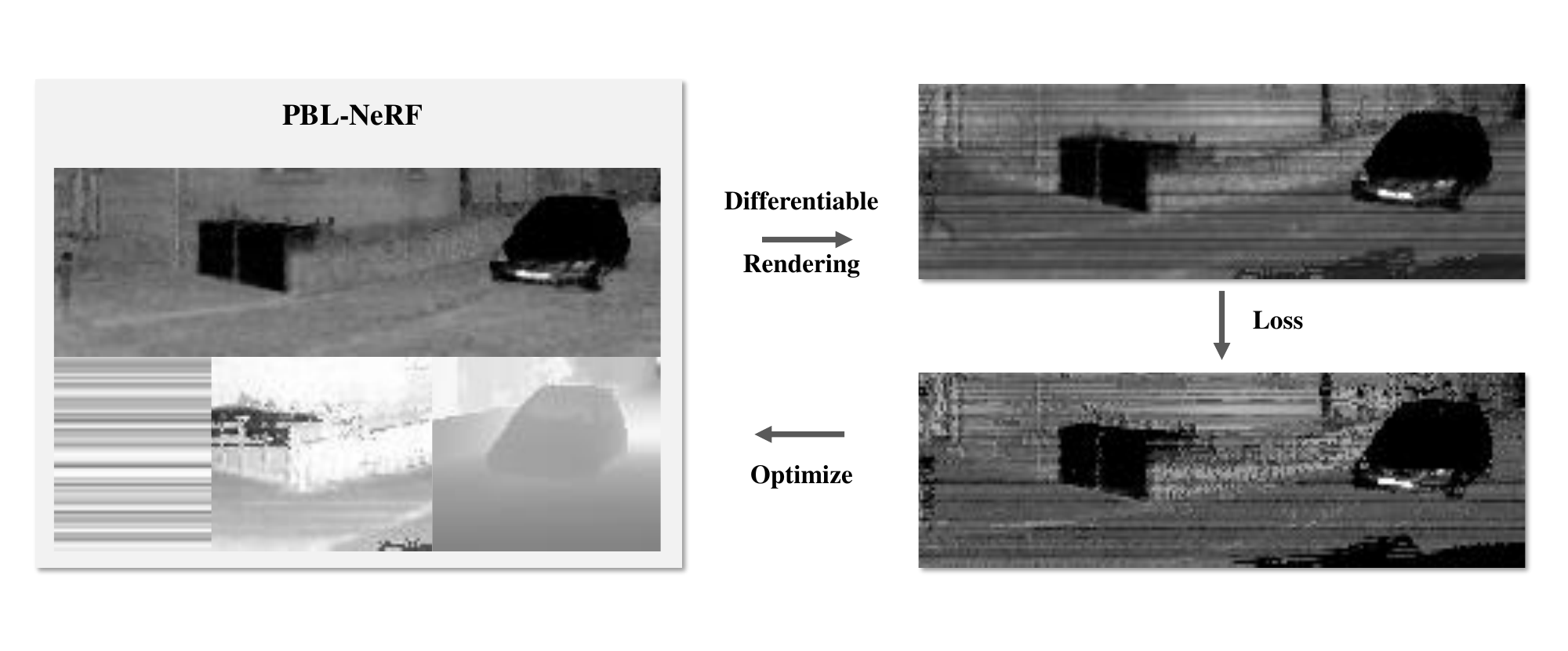} %
    \caption{PBL Pipeline: we use the direct intensity prediction as an intermediate, enhance it with physical sensor modeling (from left to right: laser power variation, cosine law and distance falloff) and optimize all parameters during training.}
    \label{fig:pipeline}
\end{figure}

NVS has become a powerful tool for realistic \li simulation where traditional techniques often fall short. By learning a dense representation of the scene, scans from arbitrary viewpoints can be synthesized. 
Significant progress has been made for supporting large-scale scenes, with primary focus on depth accuracy and handling dynamic environments. 
Yet, incomplete \li models fail to account for crucial effects such as rolling shutter, distance-dependent intensity falloff, incidence angle attenuation, and  laser beam power variation. 
The latter in particular has, to the best of our knowledge, not been addressed in the context of \li simulation but drastically affects the intensity return of the sensor, as can be seen in Fig.~\ref{fig:pipeline} and Fig.~\ref{fig:opt_params}.
This is a critical limitation since intensity information plays a vital role in many downstream applications including object detection, classification, and segmentation.

\begin{figure}[tbp]
    \centering
    \begin{subfigure}[b]{0.95\linewidth}
        \includegraphics[width=\linewidth]{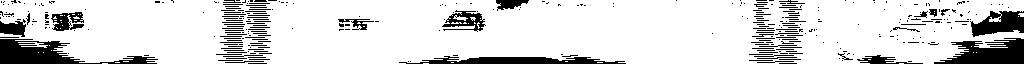}
        \end{subfigure}

        \vspace{0.2cm} 

        \begin{subfigure}[b]{0.95\linewidth}
        \includegraphics[width=\linewidth]{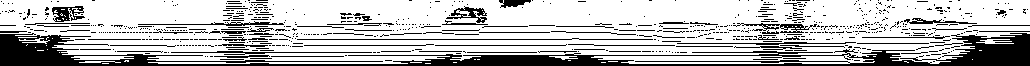}
        \end{subfigure}
        \caption{Raydrop masks resulting from improved (top) and baseline (bottom) point cloud to range view projection.}
        \label{fig:raydrop}
        \vspace{-.5em}
\end{figure}

\begin{figure*}[tbp]
    \centering
    \begin{subfigure}[c]{0.9\linewidth}
        \includegraphics[width=\linewidth]{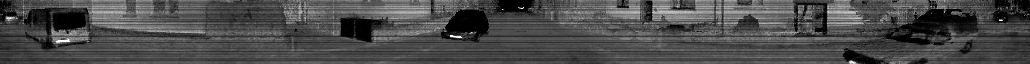} %
    \end{subfigure}

    \vspace{0.05cm} 

    \begin{subfigure}[c]{0.9\linewidth}
        \includegraphics[width=\linewidth]{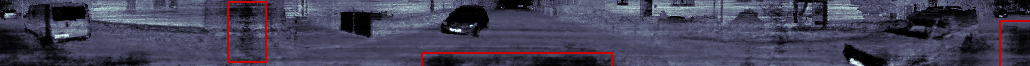} %
    \end{subfigure}
     \caption{Improved sensor modeling and masking (top) produces cleaner intensity predictions than SOTA methods~\cite{zheng_lidar4d_2024} (bottom).}
    \label{fig:intensity_masking}
      \vspace{-.5em}
\end{figure*}

\begin{figure}[tbp]
    \centering
    \begin{subfigure}[b]{0.55\linewidth}
        \centering
        \includegraphics[width=\linewidth, trim=0 0 80 0, clip]{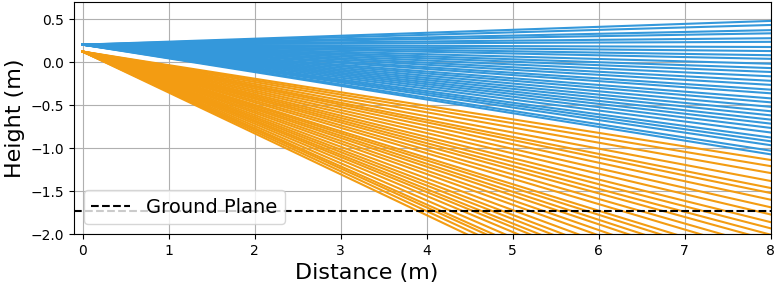}
    \end{subfigure}
    \hfill 
    \begin{subfigure}[b]{0.42\linewidth} 
        \centering
        \includegraphics[width=\linewidth]{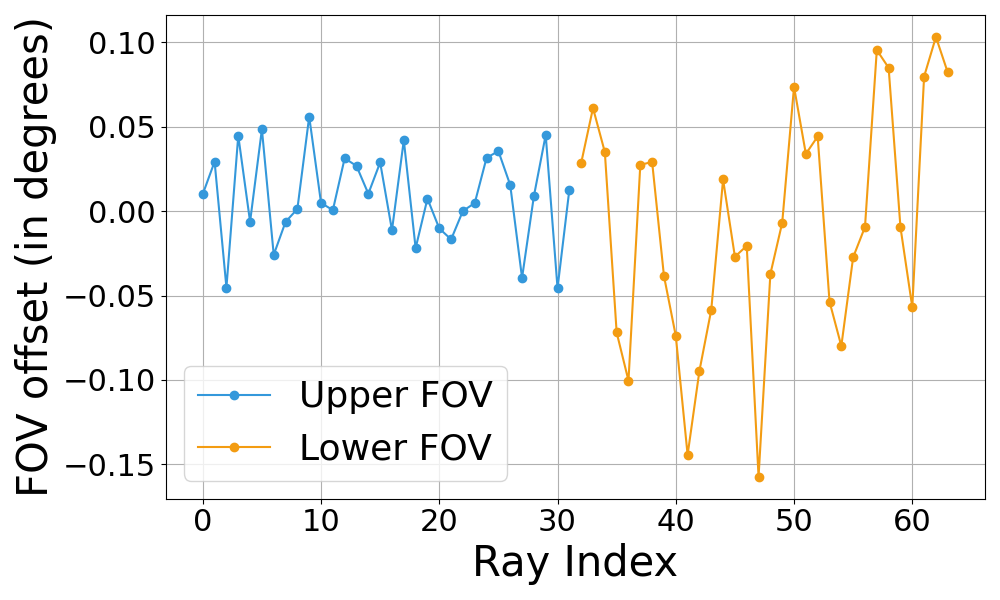}
    \end{subfigure}
    \caption{Optimized Velodyne HDL-64E intrinsics, on the left a side view of the sensor FOV and on the right the laser offsets.}
    \label{fig:optimized_intrinsics} 
    \vspace{-.5em}
\end{figure}

A further issue is data processing with incorrect sensor \emph{intrinsics}, which leads to misalignment issues in the resulting point clouds.
In this context, \li intrinsics primarily describe FOV and resolution, but can also encompass positional and rotational offsets for point cloud projection.
This is the case for the Velodyne HDL-64E sensor used in popular datasets like KITTI-360~\cite{liao_kitti-360_2023}, which uses a special configuration, compare Fig.~\ref{fig:optimized_intrinsics}.
For effective resimulation with different sensor configurations, these effects must not remain embedded within the neural representation but be explicitly modeled and separated. Our work addresses this gap in multiple ways:
\begin{itemize}
    \item We introduce enhanced sensor intrinsics to \li NVS, see Section~\ref{sec:enhanced_sensor_modeling}, which allows us to use corrected range view representations (Fig.~\ref{fig:raydrop}).
    \item We optimize the sensor parameters during training, exploiting the differentiable rendering pipeline, see Section \ref{sec:physically_based_reconstruction_enhancements} and Fig.~\ref{fig:pipeline}.
    With this, we outperform previous methods by a significant margin.
    \item We show that our approach can be used to reconstruct surface properties and generate a multimodal high resolution \li dataset in the camera perspective, see Section~\ref{sec:resimulation}. Fig. \ref{fig:intensity_masking} shows the difference between our method and previous SOTA for the intensity synthesis.
    
\end{itemize}
We call this the \emph{Physical Based LiDAR} (PBL) pipeline and base our implementation on LiDAR4D~\cite{zheng_lidar4d_2024} using the KITTI-360 dataset~\cite{liao_kitti-360_2023} for evaluation, but the same principles can be applied to other datasets and differential rendering methods.

\section{Related Work}
\subsection{\li Simulation}
As \li NVS is one special approach within the broader field of \li simulation, related approaches can offer alternatives for realistic simulation or principles we can build upon.
One direction are traditional 3D rendering techniques.
CARLA~\cite{dosovitskiy_carla_2017} comes with a basic ray tracing \li implementation but does not simulate intensity  surface interactions. 
Other methods~\cite{saleh_voxelscape_2023, marcus_synth_2025} follow a similar concept, but approximate surface reflection properties from semantic, depth, or RGB information. 
While this facilitates the generation of large scale datasets, realism is lacking.
 Data-driven approaches~\cite{guillard_learning_2022, zyrianov_learning_2022, ran_towards_2024} use real data instead to train a neural network that simulates the \li behavior.
This can either be done in an unconditional way or be conditioned with RGB or semantic information.
Methods that focus on specific \li effects complement these.
Dai et al.~\cite{dai_lidar_2022} propose intensity normalization to merge multiple \li scans into a single point cloud with consistent intensity values.
Viswanath et al.~\cite{viswanath_reflectivity_2024} includes similar effects but adopts an improved distance falloff model.
Both employ a statistical approach to estimate the parameters of the model as preprocess.
Another approach~\cite{anand_advancing_2025} directly uses different modalities as input for intensity prediction.
A related direction is the optimization of the sensor parameters.
Hu et al.~\cite{hu_rangeldm_2024} and Bewley et al.~\cite{bewley_range_2021} optimize the sensor intrinsics to improve for point cloud generation and object detection, respectively.

 Our goal is to unite these different aspects of \li simulation into a single framework that can optimize the sensor parameters through differentiable rendering. 

\subsection{Novel View Synthesis}
Novel View Synthesis (NVS) has seen rapid advances with the introduction of Neural Radiance Fields (NeRFs)~\cite{mildenhall_nerf_2020}, enabling photorealistic image synthesis through scene reconstruction from sparse observations. Recent developments such as Gaussian Splatting (GS)~\cite{kerbl_3d_2023} have further improved rendering efficiency and quality. For a comprehensive overview, we refer to the survey by Chen et al.~\cite{chen_survey_2025}.

For LiDAR simulation, NVS methods have been adapted to the unique \li specifics. Early approaches achieved high-quality synthesis without differentiable rendering~\cite{manivasagam_lidarsim_2020,li_pcgen_2023}, providing strong baselines for LiDAR scene synthesis. More recent work has focused on neural representations tailored for LiDAR, initially based on NeRFs~\cite{huang_neural_2023,tao_lidar-nerf_2024,zhang_nerf-lidar_2024, xue_geonlf_2024}.
NFL~\cite{huang_neural_2023} presented a strong basis for the physical \li modeling, incorporating multiple hits individual rays, where we focus on effects that more directly influence the intensity behavior.
With the importance of dynamic scenes in driving scenarios, great effort has been put into handling these.~\cite{wu_dynamic_2024,zheng_lidar4d_2024,tonderski_neurad_2024}.

Recently, GS methods have found application in \li settings.~\cite{hess_splatad_2025,chen_lidar-gsreal-time_2025,zhou_lidar-rt_2025}. 
SplatAD~\cite{hess_splatad_2025} introduces rolling shutter compensation by shifting the Gaussians during optimization.
LiDAR-RT~\cite{zhou_lidar-rt_2025}, on the other hand, does implement rays following 3D Gaussian Raytracing~\cite{moenne-loccoz_3d_2024} with a bounding volume hierarchy for acceleration.
It further relies on having information on dynamic objects, which results in better quality, but is not always available.
With this, our method can also be applied to more performance critical use cases.
LiDAR4D~\cite{zheng_lidar4d_2024}, which we use as the basis for our work, follows spatio-temporal approaches and operates without priors.

\section{Enhanced Sensor Modeling}
\label{sec:enhanced_sensor_modeling}
\li scans are usually represented as 3D point clouds with coordinates $(x,y,z)$ and intensity $I$.
Gaps from missing sensor returns are typically referred to as \emph{raydrop}.
For integration of \li data into the NeRF pipeline, the scans can be interpreted as panoramic images via \emph{range view} representation, where each pixel corresponds to a \li point with elevation angle $\theta$ and azimuth angle $\phi$.
While the horizontal field of view is 360 degrees, the vertical field of view is limited to $f$ with an offset $f1$ relative to the horizontal plane and $f0 = f1 -f$.
Given the point coordinates $(x,y,z)$, we can compute the \li angles $\theta$ and $\phi$ as follows:
\begin{align}
\begin{pmatrix}
\theta \\
\phi
\end{pmatrix}
&= 
\begin{pmatrix}
\arctan2(z, \sqrt{x^2 + y^2}) +{{f0}} \\
\arctan2(y, x)
\end{pmatrix}
\label{eq:angles}
\end{align}
Pixel coordinates $(i,j)$ can consequently be computed as
\begin{align}
\begin{pmatrix}
i \\
j
\end{pmatrix} &= 
\begin{pmatrix}
(1 - \theta /f) \cdot H   \\
(0.5 - \phi/2\pi) \cdot W  
\end{pmatrix}
\label{eq:pixel_coordinates}
\end{align}
where $H$ and $W$ are height and width of the range view image.
Note that $i$ also corresponds to the ring index of the \li sensor and $j$ to the timestamp within one sensor rotation.

However, this is only true when the origin of all \li rays is at (0,0,0), which is not the case for, e.g., the Velodyne HDL-64E \li sensor. 
In fact, the internal composition can be understood as two separate units stacked vertically. 
These do not only differ in their optical center but also have different FOVs, the upper one more narrow to maintain a higher resolution for distant observations.
 This is mainly an issue when generating range views from the point cloud, as the point coordinates in 3D are correct, meaning that the sensor has an internal calibration for these effects.
Hence, usual applications like object detection do not require special care, but for \li NVS it is crucial to account for this.

\subsection{Optical Center and per Diode Adjustment}
To further compensate for per laser diode deviations, we extend the elevation computation with additional offsets $\delta_i$ as well as separate FOVs $f_k$ and vertical offsets $z_k$ for each sensor unit $k$:
\begin{equation}
    \theta_k(i) = \arctan2(z-z_k, \sqrt{x^2 + y^2}) +{f0_k} + \delta_{i} \\
\end{equation}
With pixel coordinates $(i,j)$, we can compute ray directions:
\begin{equation}
\begin{pmatrix}
\theta_k(i) \\
\phi(j)
\end{pmatrix}
=
\begin{pmatrix}
(1 - \frac{i}{H}) \cdot f_k - f0_{k} - \delta_i \\
(0.5 -\frac{j}{W}) \cdot 2\pi
\end{pmatrix}
\label{eq:pixel_angles}
\end{equation}
So the range view image $P_k$ at pixel coordinates $(i,j)$ is computed as 
\begin{equation}
P_k(i,j) = (d(\theta_k(i), \phi(j)),I(\theta_k(i), \phi(j))),
\label{eq:range_image_content}
\end{equation}
where $d$ and $I$ are the measured \li distance and intensity.
\subsection{Optimization Preprocess}
Often, complete sensor intrinsics are not given, but precomputed range view images or points stored with row index and timestamp can replace this information.
Assuming that the point cloud is accurate, we can optimize the parameters of the sensor model using a reprojection loss function and the ground truth image $P$:
\begin{equation}
    \mathcal{L}_{opt} = \sum_{n}^{N} \sum_{i,j}^{H,W} \|P_n(i,j) - \hat{P_n}(i,j)\|_2^2
    \label{eq:opt_loss}
\end{equation}
Importantly, we can extend the image to $P(i,j) = (d, I, i, j)$ to include the difference between predicted and ground truth pixel coordinates within the loss.

For KITTI-360, these parameters are not available, so other methods ~\cite{bewley_range_2021, hu_rangeldm_2024} opted to reconstruct these with Hough-Voting~\cite{hough_method_1962, duda_use_1972, ballard_generalizing_1981}.
The main concept here is that good sensor intrinsics minimize the loss between the 3D reprojection the projected range view image and the original point cloud.
However, it is still possible to obtain row indices from the order of the points in the data file and their azimuth angles $\phi$ and thus directly use (\ref{eq:opt_loss}).

\subsection{Rolling Shutter}
Correct ray generation requires implementing the rolling shutter effect of spinning \li sensors.
With a frequency of 10Hz, the vehicle typically travels between one and three meters within one cycle.
We found that linear interpolation of quaternions to obtain in-between poses is sufficient for our purposes, higher order interpolation might be a more precise alternative.

\section{Physically Based Reconstruction}
\label{sec:physically_based_reconstruction_enhancements}
So far, we have obtained optimal rays for the range views and NeRF training, but there also are important interactions regarding the intensity return of the sensor.
The following part focuses on the core PBL pipeline introduced before (cf. Fig.~{\ref{fig:pipeline}}).

In camera based scene reconstruction, the images are the result of complex interactions between scene surfaces and multiple light sources, but it is usually assumed that lighting is static. 
This still leaves us with view dependent differences for reflective surfaces, which often are handled via spherical harmonics.
For \li sensors, the setup is different: there only is one light source, but it is identical to the view position and direction.
This facilitates the modeling of view dependent effects immensely but also means that the impact of including these into the simulation is greater.
We include such sensor behavior into the NeRF training pipeline by optimizing the parameters of the sensor model during training:
\begin{equation}
    \Theta^*, N_d^*, l^*  = \arg\min_{\Theta,N_d, l}  \mathcal{L}_{\text{NeRF}}(\hat{P} , \hat{N_d}, \hat{l} | P, \Pi)
\end{equation}
where $\Theta$ are the NeRF parameters, $N_d$ the distance normalization parameters, $l$ the laser powers and $\Pi$ the \li poses.

\subsection{Laser Power Normalization}
The first striking artefact in the intensity images (cf. Fig.~\ref{fig:pipeline}) is a consistent inter-row brightness variance.
The problem here is twofold: \begin{itemize}
    \item These lines are encoded into the NeRF, generating novel views from a different height will still reproduce the observed brightness differences.
    \item There is conflicting information regarding the returned intensity of surfaces depending on the laser diode of the respective ray, resulting in a blurry volume.
\end{itemize}
This pattern can be approached through parameter optimization in differential rendering, where statistical analysis fails to distinguish between causes for intensity variation.
Distance and the angle of incidence have strong influence on this. Often, on-board processing tries to compensate for this, however, this often does not work well (see later). We thus include several normalizations that allow our optimizer to account for these effects. 
We can directly include the optimizable parameters into the intensity computation with the differentiable rendering pipeline:
\begin{equation}
    I^* =  I \cdot N_d \cdot N_r \cdot l_i
    \label{eq:i}
\end{equation}

Here, the original intensity $I$ is enhanced by directly multiplying it with normalizing factors $N_d$, for the distance, $N_r$ for the incidence angle and laser powers $l_i$ with separate factors for every laser diode $i$.
In the end, this intensity will be used for the intensity loss, see Section \ref{sec:loss_computation}.
\begin{figure}[tbp]
    \begin{subfigure}[b]{0.495\linewidth}
        \includegraphics[width=\linewidth, trim=5 0 25 35, clip]{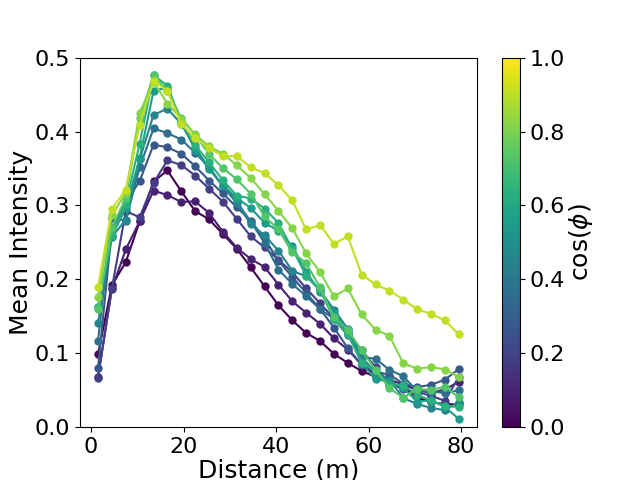} 
    \end{subfigure}
    \begin{subfigure}[b]{0.495\linewidth}
        \includegraphics[width=\linewidth, trim=5 0 25 35, clip]{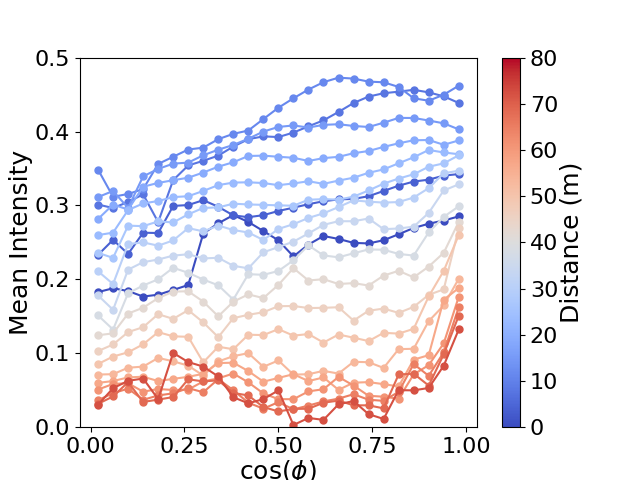} 
    \end{subfigure}
    \caption{Statistical analysis of intensity for Velodyne HDL-64E, grouped by incidence angle (left) and distance (right).}
    \label{fig:statistical_analysis}
      \vspace{-.5em}
\end{figure}

\subsection{Distance Normalization}
\label{sec:distance_normalization}
The laser used in \li sensors exhibits beam divergence~\cite{huang_neural_2023}, meaning that size of laser beams widens, leading to less energy return from distant surfaces, resulting in a quadratic falloff.

\subsubsection{Statistical Analysis}
Even though the intensity returns are largely subject to the unknown surface properties, we can get a rough estimate over the actual behavior if we analyze the intensity-distance behavior grouped by incidence angle.
We confirm two principles: the intensity generally falls over distance and for shallow incidence angles, but not necessarily following the expected trends completely, see Fig. \ref{fig:statistical_analysis}.
One factor is sensor specific internal preprocessing or normalization to output meaningful values for perception tasks, but more complex physical interactions also play a role in this.
Still, our general solution can be applied analog to the laser power, i.e., we optimize  depth and incidence normalization parameters during training that incorporate this.

\subsubsection{Distance Falloff}
\label{sec:distance_falloff}
To constrain the search space, we choose a function that balances the observed and the theoretical behavior.
At its core, we still use the quadratic falloff model $f = \frac{1}{\Delta^2}$ to account for light falloff and beam divergence.
To model the near range effect, we shift by distance $d_{near}$ and add a scale factor $s$, resulting in:
\begin{equation}
    \Delta_d= (s(d - d_{near})+1)
    \label{eq:dfar_base}
\end{equation}
We enable non-quadratic falloffs by dividing the falloff exponent by $q$ and normalizing the function, arriving at the final equation:
\begin{equation}
    D_\text{far} = {\Delta_d}^{-\frac{2}{q}} \cdot q - q + 1
    \label{eq:dfar}
\end{equation}
Here, $q$ is not a parameter that captures a specific physical light interaction but rather allows our approach to be flexible for sensor internal processing differences.

\begin{figure}[tbp]
    \centering
    \begin{subfigure}[b]{0.417\linewidth}
        \includegraphics[width=\linewidth, trim=0 0 0 0, clip]{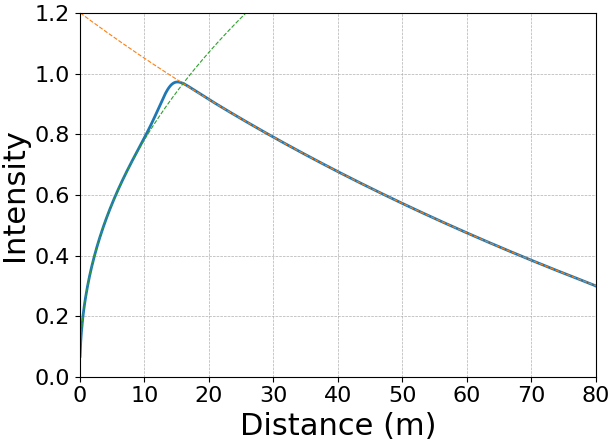} 
    \end{subfigure}
    \begin{subfigure}[b]{0.57\linewidth}
        \includegraphics[width=\linewidth, trim=0 5 0 0, clip]{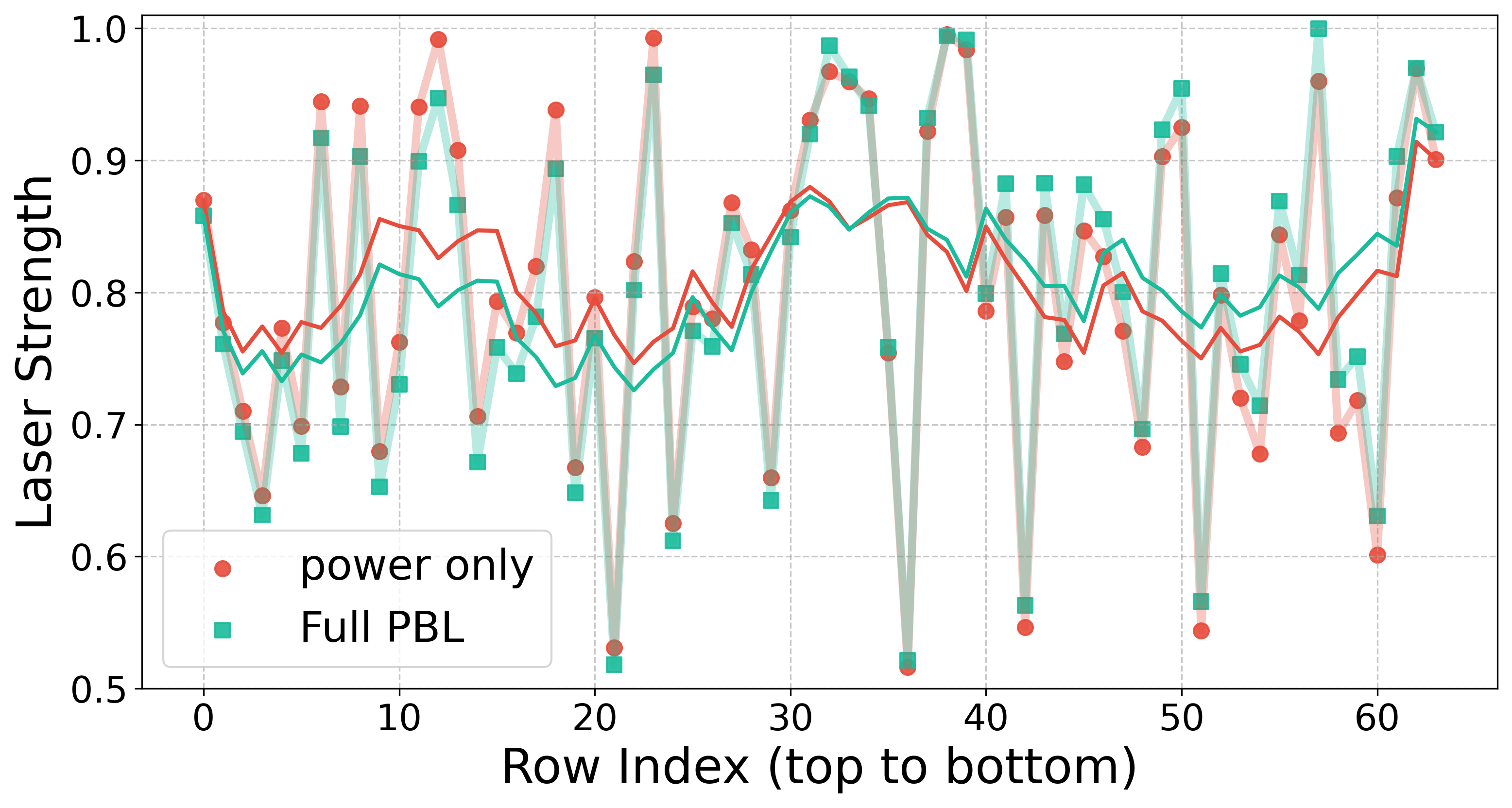} 
    \end{subfigure}
    \caption{Distance function and laser power after optimization.}
    \label{fig:opt_params}
      \vspace{-.5em}
\end{figure}

In contrast, the intensity decrease in proximity until range $d_\text{near}$ originates from lens defocusing~\cite{biavati_correction_2011}.
Viswanath et al.~\cite{viswanath_reflectivity_2024} adopt the modeling from Fang et al.~\cite{fang_intensity_2015}:
\begin{equation}
    \eta(d) = 1 - \exp\left\{ \frac{-2  r_d^2  (d+\delta)^2}{D^2  S^2}\right\}  
    \label{eq:eta}
\end{equation}
with $r_d$ as laser detector radius, $\delta$ as offset between measured and actual distance, $D$ as lens diameter, and $S$ as focal length. 
For optimization during NeRF training, we can replace the sensor parameters with $s_\eta= \frac{2r_d^2}{D^2S^2}$, where $\delta$ remains as an offset parameter.
\begin{equation}
    \eta(d) = 1 - e^  {- s_\eta  (d+\delta)^2 }
\end{equation} 	
Still it is not quite ideal for our use case, where we want to retain more intensity nearing 0, so we opted for 
a fractional power falloff with optimizable parameters $s_N$ and $q_n$ analog to $q$ and $s$ for the far range distance falloff in (\ref{eq:dfar_base}) and (\ref{eq:dfar}).
\begin{equation}
    \eta(d) =  s_\eta d^{1/q_\eta}, q_n > 0
\end{equation}
We interpolate between the near range and distance falloff using the sigmoid function with offset $d_\text{near}$ and steepness $k$.
\begin{equation}
    \sigma(d) = \frac{1}{1 + e^{-k(d - d_{near})}}   
\end{equation}
 The final distance falloff function is then defined as:
\begin{equation}
   N_d= \sigma \cdot D_{far} + (1 - \sigma)\cdot \eta
\end{equation}
We observe that the system compensates for correlations between effects, for example, depending on the elevation angle, certain distances or incidence angles are more likely, compare the optimized laser power factors in Fig.~\ref{fig:opt_params}. Optimizing only the laser power wrongly attributes incidence and distance attenuation to the laser power.
\subsection{Incidence Normalization}
\label{sec:incidence_normalization}
Our starting point here is that, for diffuse surfaces, the intensity return depends on the incidence $\phi_n$ between the surface normal and the sensor.
If we look back to the statistical data in Fig. \ref{fig:statistical_analysis}, we can see that there is great variety for this in practice.
For other \li sensors, the measured data follows the cosine law more closely~\cite{dai_lidar_2022}, so we can assume sensor processing has an influence here again.
Our approach accounts for this with a reflectivity coefficient $r$  that serves as an additional modality of the NeRF.
This way, we do not rely on a global model of the surface reflectivity but can reconstruct different surface properties within the neural scene representation.

\subsubsection{Incidence Angle Computation}
We can adopt our previously defined angle computation from (\ref{eq:pixel_angles}) to compute the incidence angle $\phi_n$ with respect to the surface normal $n$.
For estimating the surface normals, we follow the approach of Schreck et al.~\cite{schreck_height_2023} and apply the Scharr filter to a 3-channel range view image of 3D coordinates $(x,y,z)$.
There are two issues with the results. First, strong object edges are very thick.
Second, the variance in laser power also leads to barely visually discernable errors in the distance measurements, which however cause incorrect horizontal edges.

\begin{figure}[tbp]
    \centering
    \includegraphics[width=0.99\linewidth]{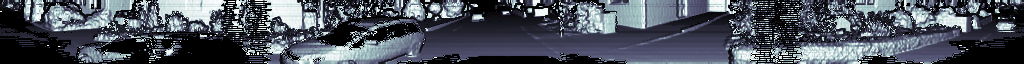} 
    \caption{Incidence panorama image, computed via $cos(\phi_n)$.}
    \label{fig:incidence_angles}
      \vspace{-.5em}
\end{figure}

An alternative to this is covariance analysis via neighboring points, which however produces more noisy results due sparsely sampled regions in individual scans.
Since we already know where strong edges are, we can define a threshold and replace respective normals accordingly.
For the horizontal edges, we use the average neighborhood normal, since the area where these edges are harmful for our pipeline are mostly flat surfaces, a final  example result is shown in Fig. \ref{fig:incidence_angles}.
\subsubsection{Reflectivity Prediction}
Based on incidence angle $\phi_n$ and surface reflectivity $s(R)$, we can formulate a normalization factor $N_R$ for the corrected intensity $I^*$ defined in (\ref{eq:i}):
\begin{equation}
   N_R = \cos(\phi_n)^{s(R)}
\end{equation}
This requires that we add an MLP that is otherwise identical to the intensity and raydrop MLPs of LiDAR4D~\cite{zheng_lidar4d_2024} using temporal aggregated planar and hash features $\text{f}_\text{planar}$ and $\text{f}_\text{hash}$.
The new MLP outputs a reflectivity $r$, using the pointwise reflectivity output $r_i$.
\begin{equation}
r = \text{MLP}_\text{reflectivity}(\text{f}_\text{planar}, \text{f}_\text{hash}, \gamma(d))
\end{equation}
Further following LiDAR4D, the accumulated transmittance $T_i$ and the opacity $\alpha_i = 1 - e^{-\sigma_i \delta_i}$ obtained from the density $\sigma$ and the distance $\delta_i$ between samples, the NeRF can predict a reflectivity image $\hat{R}$:
\begin{equation}
    \hat{R} = \sum_{i=1}^{N} T_i \cdot \alpha_i \cdot r_i
\end{equation}
As higher reflectivity values quickly lead to extreme falloffs, we get higher numerical precision by defining $s(\hat{R})$ with empirically determined parameters $a$ and $b$:
\begin{equation}
    s(\hat{R}) = (a \cdot \hat{R})^{b}, \quad \hat{R} \in [0,1]
\end{equation}
We use a target reflectivity value $r_t$ to constrain the reflectivity parameter by adding a loss that punishes low reflectivity:
\begin{equation}
    \mathcal{L}_{R} = \text{ReLU}(r_t - \text{median}(\hat{R}))
\end{equation}
This prevents the overall system from letting the reflectivity mostly converge towards maximum diffusity and reflectivity, in which case changes in surface reflectivity would again be only captured by the direct intensity output.

\subsection{Loss Computation}
\label{sec:loss_computation}
We can now combine all of our sensor effects for $I^*$ to formulate the overall loss function $\mathcal{L}$.
The depth and raydrop losses $\mathcal{L}_{D} $ and $\mathcal{L}_{M}$ are same as in LiDAR4D~\cite{zheng_lidar4d_2024}.
For the intensity however, we insert our enhanced model $I^*$:
\begin{equation}
    \mathcal{L}_{I^*} = \frac{1}{N} \sum_{i=1}^{N} \|I^*_i - \hat{I_i}\|_2^2
\end{equation}
Adding further regularization of the laser power via
\begin{equation}
    \mathcal{L}_{laser} = \frac{1}{N} \sum_{i=1}^{N} \text{ReLU}(l_i - 1.0)
\end{equation}
 we arrive at the final enhanced loss $\mathcal{L}$:
\begin{equation}
    \mathcal{L} = \lambda_\alpha \cdot \mathcal{L}_{D} + \lambda_\beta \cdot \mathcal{L}_{I^*} + \lambda_\gamma \cdot \mathcal{L}_{M} +
    \lambda_r \cdot \mathcal{L}_{R} + \lambda_l \cdot \mathcal{L}_{l}
\end{equation}
Raydrop (cf. Fig. \ref{fig:raydrop}) serves two different purposes: predicting areas, where there is no \li return, but also masking out invalid pixels during training such as self occlusions.
As with our improved projection, we prevent gaps between the laser rows, we can emphasize the former use of raydrop.
In particular, we build a statistical global drop mask for self occlusions that is used for masking the raydrop loss, so in our pipeline the raydrop only predicts the sensor behavior for surfaces.

For the intensity, we observe a further KITTI-360 specific issue: there are significant artefacts horizontally around +/-45 degrees, where the depth is correct when the measurement is successful, but the intensity is 0.
We again iterate over all images and add consistently affected areas to a global intensity mask.

When employing incidence angle normalization, the results depend on the quality of the normals.
In particular, very shallow, incorrect incidence angles interfere with the reflectivity prediction, so we define a threshold to further mask the intensity loss computation in these cases.
We will evaluate the effects of these among others in the following.
\begin{table*}[tbp]
    \centering
    \caption{Comparison between previous methods (top) and our method (bottom), highlighting \colorbox{mygreen}{best} and \colorbox{myyellow}{second best}. LiDAR4D~\cite{noauthor_ispc-lablidar4d_2025} reports improved results compared to the original paper~\cite{zheng_lidar4d_2024}.}
    \setlength{\tabcolsep}{5.5pt}  
    \begin{tabular}{l|lllll|lllll|ll}
    \hline
    & \multicolumn{5}{c|}{Intensity} & \multicolumn{5}{c|}{Depth} & \multicolumn{2}{c}{Point Cloud} \\
    Method & RMSE$\downarrow$ & MedAE$\downarrow$ & LPIPS$\downarrow$ & SSIM$\uparrow$ & PSNR$\uparrow$ & RMSE$\downarrow$ & MedAE$\downarrow$ & LPIPS$\downarrow$ & SSIM$\uparrow$ & PSNR$\uparrow$ & CD$\downarrow$ & $F_1\uparrow$ \\ \hline
    PCGen~\cite{li_pcgen_2023}                 & 0.1970 & 0.0763 & 0.5926 & 0.1351 & 14.1181 & 5.6853 & 0.2040 & 0.5391 & 0.4903 & 23.1675 & 0.4636 & 0.8023 \\
    LiDARsim~\cite{manivasagam_lidarsim_2020}  & 0.1666 & 0.0569 & 0.3276 & 0.3502 & 15.5853 & 6.9153 & 0.1279 & 0.2926 & 0.6342 & 21.4608 & 3.2228 & 0.7157 \\
    LiDAR-NeRF~\cite{tao_lidar-nerf_2024}      & 0.1464 & 0.0438 & 0.3590 & 0.3567 & 16.7621 & 4.0886 & 0.0556 & 0.2712 & 0.6309 & 26.0590 & 0.1502 & 0.9073 \\
    LiDAR4D~\cite{zheng_lidar4d_2024}          & 0.1195 & 0.0327 & 0.1845 & 0.5304 & 18.5561 & 3.5256 & 0.0404 & 0.1051 & 0.7647 & 27.4767 & 0.1089 & 0.9272 \\
    LiDAR4D~\cite{noauthor_ispc-lablidar4d_2025} & \cellcolor{myyellow}0.1043 & 0.0297 & \cellcolor{mygreen}\textbf{0.1618} & \cellcolor{myyellow}0.6373 & \cellcolor{myyellow}19.6480 & \cellcolor{myyellow}2.9427 & \cellcolor{myyellow}0.0315 & \cellcolor{myyellow}0.0811 & \cellcolor{myyellow}0.8656 & \cellcolor{myyellow}28.9969 & \cellcolor{myyellow}0.0935 & \cellcolor{myyellow}0.9317 \\                                 
    LiDAR-RT~\cite{zhou_lidar-rt_2025}         & 0.1115 & \cellcolor{myyellow}0.0271 & \cellcolor{myyellow}0.1812 & 0.6077 & 19.0862 & 3.4671 & 0.0512 & 0.1016 & 0.8406 & 27.6755 & 0.1077 & 0.9255 \\
    \hline
    LiDAR4D++                                  & 0.0939 & 0.0355 & 0.2126 & 0.5612 & 20.5966 & 1.1539 & 0.0287 & 0.0279 & 0.9791 & 37.1387 & 0.0618 & 0.9569 \\
    ~+ Rolling Shutter   & 0.0936 & 0.0355 & 0.2122 & 0.5621 & 20.6317 & 1.1329 & 0.0282 & 0.0277 & 0.9792 & 37.2916 & 0.0615 & 0.9583 \\
    ~+ PBL & \cellcolor{mygreen}\textbf{0.0868} & \cellcolor{mygreen}\textbf{0.0245} & 0.1847 & \cellcolor{mygreen}\textbf{0.6932} & \cellcolor{mygreen}\textbf{21.2595} & \cellcolor{mygreen}\textbf{1.1281} & \cellcolor{mygreen}\textbf{0.0281} & \cellcolor{mygreen}\textbf{0.0264} & \cellcolor{mygreen}\textbf{0.9794} & \cellcolor{mygreen}\textbf{37.3365} & \cellcolor{mygreen}\textbf{0.0574} & \cellcolor{mygreen}\textbf{0.9594} \\ \hline
    \end{tabular}
    \label{tab:quantitative_comparison}
\end{table*}

\section{Evaluation}
\begin{figure}[bp]
    \centering
    \newsavebox{\imagebox}
    \sbox{\imagebox}{\includegraphics{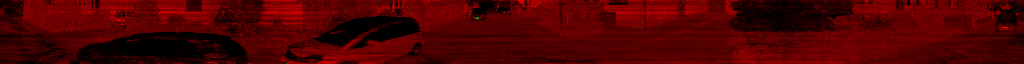}}
    \begin{subfigure}[b]{00.9\linewidth}
        \includegraphics[width=\linewidth, clip, viewport={0 0 {0.5\wd\imagebox} {\ht\imagebox}}]{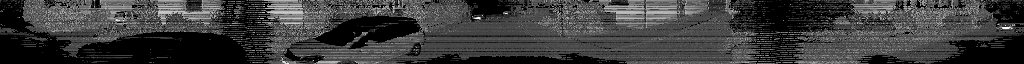}
        \caption{Ground Truth Intensity}
    \end{subfigure}
    \begin{subfigure}[b]{00.9\linewidth}
        \includegraphics[width=\linewidth, clip, viewport={0 0 {0.5\wd\imagebox} {\ht\imagebox}}]{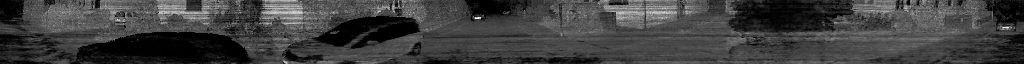}
        \caption{LiDAR4D++: Intensity Prediction ($I$)}
    \end{subfigure}
    \begin{subfigure}[b]{00.9\linewidth}
        \includegraphics[width=\linewidth, clip, viewport={0 0 {0.5\wd\imagebox} {\ht\imagebox}}]{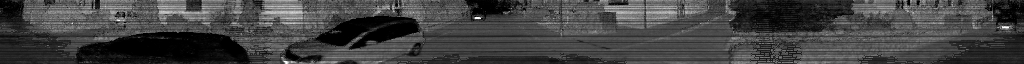}
        \caption{PBL: Enhanced Intensity Prediction ($I^*$)}
    \end{subfigure}
    \begin{subfigure}[b]{00.9\linewidth}
        \includegraphics[width=\linewidth, clip, viewport={0 0 {0.5\wd\imagebox} {\ht\imagebox}}]{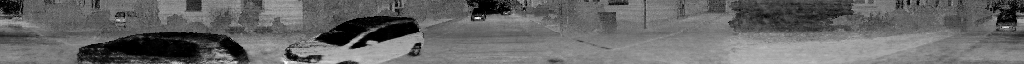}
        \caption{PBL: Base Intensity Prediction ($I$)}
    \end{subfigure}
    \begin{subfigure}[b]{00.9\linewidth}
        \includegraphics[width=\linewidth, clip, viewport={0 0 {0.5\wd\imagebox} {\ht\imagebox}}]{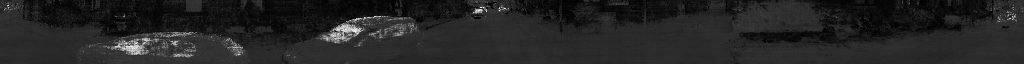}
        \caption{PBL: Reflectivity Prediction ($R$)}
    \end{subfigure}
    \caption{The PBL pipeline generates the final intensity image $I^*$ by reconstructing the base intensity $I$, reflectivity $R$ and sensor effects, see (\ref{eq:i}).}
    \label{fig:qualitative_comparison}
\end{figure}

\subsection{Comparison with Previous Work}
We compare our method against SOTA methods on KITTI-360~\cite{liao_kitti-360_2023}.
We follow the protocol of LiDAR4D~\cite{zheng_lidar4d_2024} using six dynamic and four static scenes with 50 to 64 frames, reserving four evenly spaced frames to evaluate how well these views can be synthesized.
With the correct projection, we can now effectively use a resolution of 1024 x 64 for training.
The metrics used for intensity and depth are Root Mean Square Error~(RMSE), Median Absolute Error~(MedAE), LPIPS~\cite{zhang_unreasonable_2018}, SSIM~\cite{wang_image_2004} and Peak Signal-to-Noise Ratio~(PSNR).
Chamfer distance (CD)~\cite{fan_point_2017} and F-Score~($F_1$) 
  allow further insights on the resulting point cloud quality.

We make one important modification to the general loss computation by always using the ground truth mask when available.
Although it is not always explicitly stated, the majority of previous approaches use the predicted raydrop mask.
This makes perfect sense for evaluating the complete NVS capabilities, but for judging the quality of the individual modalities the ground truth mask is more appropriate.
This further allows to use the raydrop mask for the actual ray surface interactions as outlined in Section~\ref{sec:loss_computation}.
Training LiDAR4D~\cite{noauthor_ispc-lablidar4d_2025} with this modification and the improved range view projection (LiDAR4D++) shows significant improvement for depth and point cloud metrics, see Tab.~\ref{tab:quantitative_comparison}.
While there are limited changes to the scene reconstruction yet, the precise loss masking improves the metrics in uncertain areas.
Conversely, the corrected range view projection initially causes larger errors, which can be attributed to the different FOV, where the image now actually contains more resolution in the distance and thus is more challenging to reproduce, refer back to Fig.~\ref{fig:intensity_masking}.

The more important comparison is the relative improvement of our PBL pipeline over the LiDAR4D++ baseline.
Simulating the rolling shutter effect during training further improves this, plausibly having a smaller but still positive impact on the intensity.
Adding PBL enhancements, our approach surpasses previous SOTA in all metrics except LPIPS.
The reason for this outlier could be that masking is problematic for the LPIPS metric; applying the incorrect \emph{striped} masks inadvertently raises similarity.
PBL overall produces sharper and more detailed intensity images compared to LiDAR4D++, see Fig.~\ref{fig:qualitative_comparison}.

\begin{table}[tbp]
    \centering
    \caption{Adding individual components to the LiDAR4D++ baseline (top) and the full PBL pipeline (bottom), reflectivity target $r_t$ as introduced in Section \ref{sec:incidence_normalization}.}
    \setlength{\tabcolsep}{5.5pt}  
    \begin{tabular}{l|lllll}
    \hline
        & \multicolumn{5}{c}{Intensity} \\
        Setting & RMSE$\downarrow$ & MedAE$\downarrow$ & LPIPS$\downarrow$ & SSIM$\uparrow$ & PSNR$\uparrow$ \\ \hline
        LiDAR4D++  & 0.0939 & 0.0355 & 0.2126 & 0.5612 & 20.5966 \\
        +Distance ($N_d$)  & 0.0918 & 0.0348 & 0.2137 & 0.5656 & 20.7794 \\
        +Intensity Mask    & 0.0969 & 0.0322 & 0.2268 & 0.5765 & 20.3216 \\
        +$N_r, r_t=0.2$    & 0.0906 & 0.0326 & 0.1901 & 0.5856 & 20.9006 \\
        +$N_r, r_t=0.0$    & 0.0888 & 0.0310 & 0.1899 & 0.6057 & 21.0729 \\
        +Laser  ($l_i$)    & 0.0882 & 0.0304 & 0.1914 & 0.6618 & 21.1372 \\
        \hline
        +PBL($r_t=0.2$) & 0.0882 & 0.0253 & \textbf{0.1814} & 0.6859 & 21.1261 \\
        +PBL($r_t=0.0$)  & \textbf{0.0868} & \textbf{0.0245} & 0.1847 & \textbf{0.6932} & \textbf{21.2595} \\
        \hline
    \end{tabular}
    \label{tab:ablation_study}
\end{table}
\begin{figure}[bp]
    \centering
    \vspace{-0.5em}
\begin{tikzpicture}[spy using outlines={rectangle, magnification=3, size=1cm, connect spies, red!80!black}]
    \node[anchor=north west] (baseline) at (0,0) {
    \begin{minipage}{0.9\linewidth}
        \includegraphicsright[width=\linewidth]{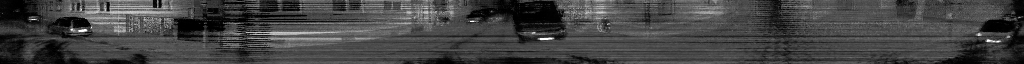}
    \end{minipage}
    };
    \spy on ($(baseline.north west)+(1.5cm, -0.5 cm)$) in node [left] at ($(baseline)+(0.6cm,0)$);
    
    \node[anchor=north west] (motion) at (0,-1cm) {
    \begin{minipage}{0.9\linewidth}
        \includegraphicsright[width=\linewidth]{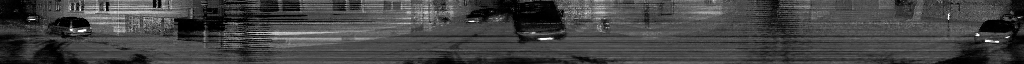}
    \end{minipage}
    };
    \spy on ($(motion.north west)+(1.5cm, -0.5 cm)$) in node [left] at ($(motion)+(0.6cm,0)$);
    
    \node[anchor=north west] (poses) at (0,-2cm) {
    \begin{minipage}{0.9\linewidth}
        \includegraphicsright[width=\linewidth]{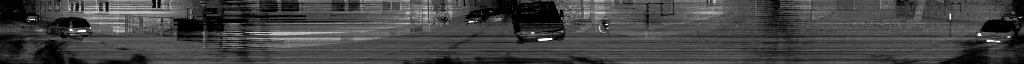}
    \end{minipage}
    };
    \spy on ($(poses.north west)+(1.5cm, -0.5 cm)$) in node [left] at ($(poses)+(0.6cm,0)$);  
\end{tikzpicture}
\caption{Geometric reconstruction quality: top: LiDAR4D++, middle: with rolling Shutter, bottom: with pose optimization.}
\label{fig:motion_compensation}
\end{figure}

\begin{figure*}[tbp]
    \centering
    \begin{subfigure}[b]{0.48\linewidth}
        \includegraphics[width=\linewidth]{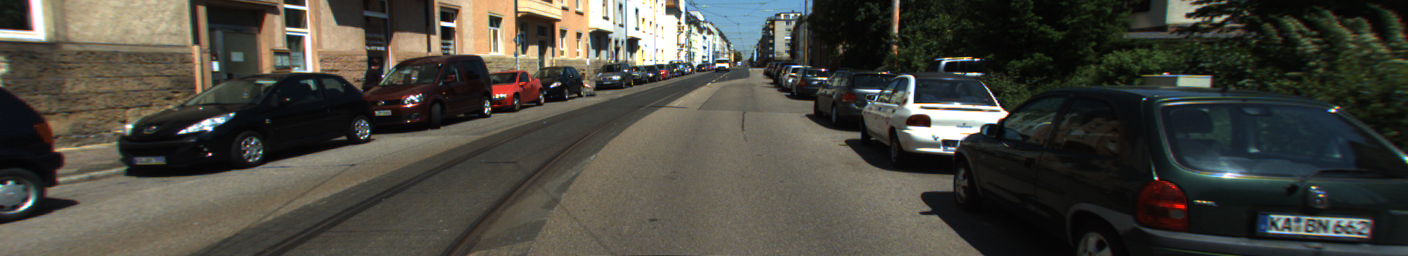}
    \end{subfigure}%
    \hspace{0.01em}
    \begin{subfigure}[b]{0.48\linewidth}
        \includegraphics[width=\linewidth]{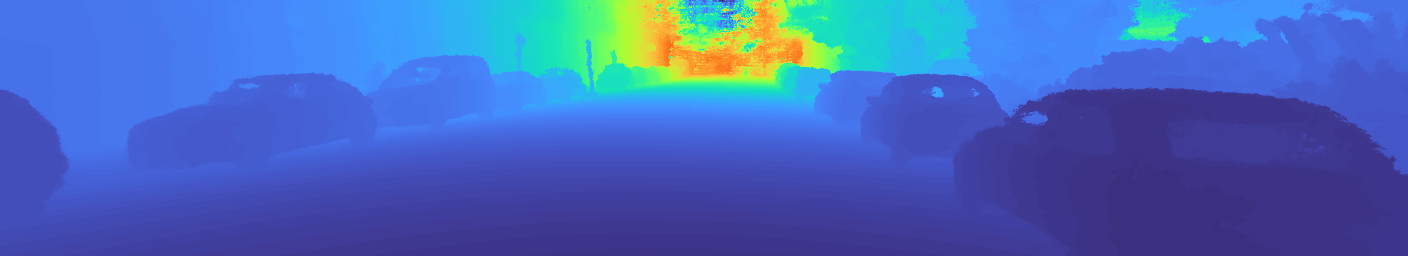}
    \end{subfigure}

    \vspace{0.24em}
    
    \begin{subfigure}[b]{0.48\linewidth}
        \includegraphics[width=\linewidth]{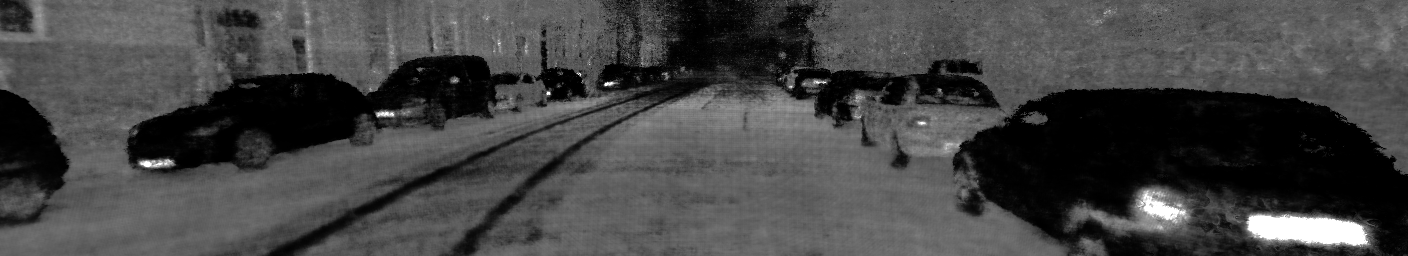}
    \end{subfigure}%
    \hspace{0.01em}
    \begin{subfigure}[b]{0.48\linewidth}
        \includegraphics[width=\linewidth]{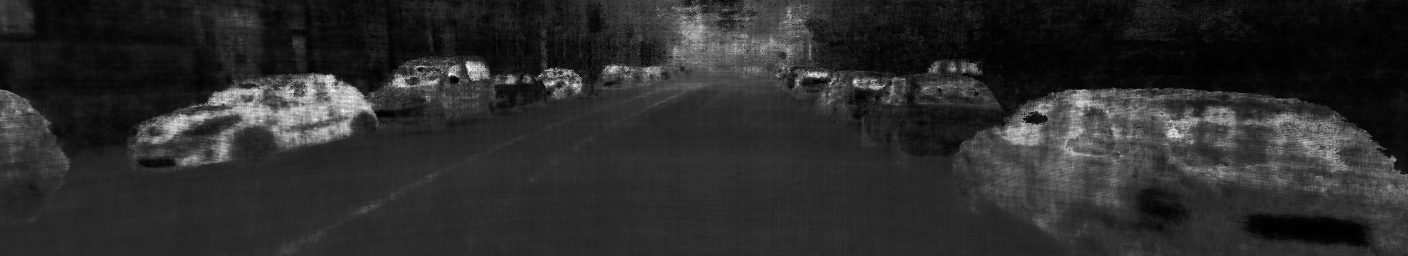}
    \end{subfigure}
    \caption{Cross-sensor resimulation showing scene from a camera viewpoint, enabling multi-modal data generation. The model successfully synthesizes high resolution LiDAR intensity, reflectivity properties and depth information.}
    \label{fig:camera_resimulation}
      \vspace{-.5em}
\end{figure*}

\subsection{Evaluating Sensor Effects}
Since the PBL pipeline consists of multiple components, we conduct ablation studies to examine individual contributions.
\subsubsection{PBL Components}
To this end, our focus lies on the intensity metrics, which have been the main target of the physically based modeling.
We observe, that every component aids in improving the intensity prediction, but the most significant improvements come from the reflectivity normalization $N_r$ and the laser power optimization.
With $r_t=0.0$, effectively no constraining happens, so the reflectivity acts more like an enhanced raydrop prediction.
Although this mostly produces better results in the metrics, higher reflectivity should often be more realistic but is difficult to infer from the limited observations.

The complete PBL pipeline, in particular through distance normalization and higher reflectivity constraining, allows recovery of intensity values in perspectives, where these areas actually do not return a signal to the sensor, (cf. Fig. \ref{fig:qualitative_comparison}). The base intensity $I$ retains information about the surface of the left vehicle.
This is exactly our desired behavior, as $I$ now acts as distance and angle independent \li \emph{color}.
When resimulating this view, the PBL components then suppress view-dependent areas to arrive at the correct intensity values.
\subsubsection{Motion Optimization}
\label{sec:motion_compensation}
We have shown that simulation results are better when training with pose-based rolling shutter approximation in Tab.~\ref{tab:quantitative_comparison}.
Yet, we could further profit from the differentiable rendering by combining this with pose optimization.
In theory, it would be possible to optimize a pose for up to every point timestamp.
We have tested this concept by starting with a very naive optimization approach and adding optimizable rotation and position offsets for the original poses, which does fix actual pose outliers, as shown in Fig. \ref{fig:motion_compensation}.
Here, the rolling shutter effect can produce a better reconstruction of the lantern on the right, but only the pose optimization is able to fix the overall misalignment.
Despite this, performance deteriorates on average, as the original poses are already mostly accurate, so more
constrained methods~\cite{xue_geonlf_2024} might be suitable for this task.

\section{Advanced Resimulation}
\label{sec:resimulation}
\begin{figure}[bp]
    \centering
    \vspace{-.5em}
    \sbox{\imagebox}{\includegraphics{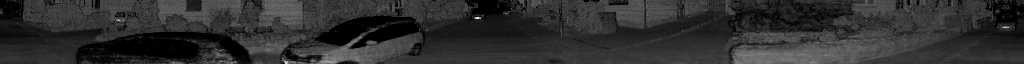}}
    \begin{subfigure}[b]{00.95\linewidth}
        \includegraphics[width=\linewidth, clip, viewport={0 0 {0.5\wd\imagebox} {\ht\imagebox}}]{images/hd/2combined.png}
    \end{subfigure}

    \vspace{0.2em}

    \begin{subfigure}[b]{00.95\linewidth}
        \includegraphics[width=\linewidth, clip, viewport={0 0 {0.5\wd\imagebox} {\ht\imagebox}}]{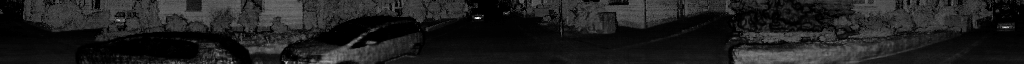}
    \end{subfigure}
    \caption{By adjusting the reflectance parameter (top: $a_r=2$, bottom: $a_r =6$), we can simulate different surface materials and lighting conditions. We simulate without distance falloff and laser power variation to produce a clean image.}
    \label{fig:reflectance_simulation}
\end{figure}

The separation of the different modalities offers strong advantages for resimulation, as we can recombine them as needed.
For simulation of other \li sensors, different distance normalization parameters or laser powers can be used, while for downstream tasks, varying the reflectivity can be useful.
One variant would be to use segmentation information to provide completely different material properties. Another possibility is to scale the reconstructed reflectivity with a factor $a_r$, so the reflectivity factor becomes $N_R(a_r) = cos(\phi_n)^{a_r s(R)}$, compare Fig.~\ref{fig:reflectance_simulation}.

While \li data has been used in neural rendering approaches to support camera NVS, we introduce a novel variant, where we use the NeRF to upsample the \li data to a camera view, thereby producing high resolution intensity, reflectivity and depth reconstruction that is perfectly aligned with the RGB images, see Fig. \ref{fig:camera_resimulation}.
Importantly, there are no occlusions effect that usually arise from projecting into a different sensor origin and through simulating the rolling shutter effect during training, we can choose to disable it for rendering novel views.
This way, we construct a multimodal dataset from all KITTI-360~\cite{liao_kitti-360_2023} sequences by training on 236 scenes with 50 frames each, resulting in more than 10k frames.
These are selected by ensuring that there are valid poses and camera segmentation masks available and at least 200 frames between two consecutive sequences, which could be used to test downstream tasks.

\section{Limitations}
\label{sec:limitations}
One issue is the applicability of our method to other datasets.
Even though we have taken care to design the optimization part so that it can fit to different sensor behaviors, manual tuning could be required depending on the specific type.
The impact of the individual sensor effects can also vary and, when the accurate intrinsics are already available, the preprocessing step can be skipped.
Furthermore, we did not implement a full multi-sensor solution but only support up to two sensors with individual vertical offset and FOV. In principle, any ray pattern can be simulated. However, for intrinsic parameter optimization, both pixel and world coordinates must be obtainable — a requirement that may not be met by all sensor models.

A more conceptual reconstruction limitation arises from the incompleteness of the training scene.
The way the optimization works, the main target is reproducibility from the given views.
Using only a limited number observations of each surface will then not result in the correct physical parameters but the simplest ones that explain these observations.
As shown in Fig.~\ref{fig:opt_params}, adding more components to the model will help, but there still is no guarantee that they will not compensate for each other, e.g. a fraction of the reflectivity properties will instead be captured within the intensity.
To prevent this, one could either collect denser data for the specific purpose of reconstructing scene properties or utilize strong physical constraints when more information for the sensor behavior is known.
\section{Conclusion}
We have proposed a \li intrinsics calibration scheme and physically based \li model (PBL) to improve neural scene reconstruction.
We hope that especially the former can find application beyond NVS, as many vision centric tasks build on \li range view representations. 

The resulting PBL pipeline -- modeling distance, incidence and laser effects -- has not only achieved significant quality improvements for NeRF-based \li synthesis, but also demonstrated the potential of neural rendering approaches for reconstruction of sensor and scene properties.

\clearpage\newpage

\bibliographystyle{IEEEtran}
\bibliography{references,l4d}

\begin{thebibliography}{10}
\providecommand{\url}[1]{#1}
\csname url@samestyle\endcsname
\providecommand{\newblock}{\relax}
\providecommand{\bibinfo}[2]{#2}
\providecommand{\BIBentrySTDinterwordspacing}{\spaceskip=0pt\relax}
\providecommand{\BIBentryALTinterwordstretchfactor}{4}
\providecommand{\BIBentryALTinterwordspacing}{\spaceskip=\fontdimen2\font plus
\BIBentryALTinterwordstretchfactor\fontdimen3\font minus \fontdimen4\font\relax}
\providecommand{\BIBforeignlanguage}[2]{{%
\expandafter\ifx\csname l@#1\endcsname\relax
\typeout{** WARNING: IEEEtran.bst: No hyphenation pattern has been}%
\typeout{** loaded for the language `#1'. Using the pattern for}%
\typeout{** the default language instead.}%
\else
\language=\csname l@#1\endcsname
\fi
#2}}
\providecommand{\BIBdecl}{\relax}
\BIBdecl

\bibitem{liao_kitti-360_2023}
Y.~Liao, J.~Xie, and A.~Geiger, ``{KITTI}-360: {A} {Novel} {Dataset} and {Benchmarks} for {Urban} {Scene} {Understanding} in {2D} and {3D},'' \emph{IEEE Transactions on Pattern Analysis and Machine Intelligence}, vol.~45, no.~3, pp. 3292--3310, 2023.

\bibitem{zheng_lidar4d_2024}
Z.~Zheng, F.~Lu, W.~Xue, G.~Chen, and C.~Jiang, ``Lidar4d: Dynamic neural fields for novel space-time view lidar synthesis,'' in \emph{{IEEE/CVF} Conference on Computer Vision and Pattern Recognition, {CVPR} 2024, Seattle, WA, USA, June 16-22, 2024}.\hskip 1em plus 0.5em minus 0.4em\relax {IEEE}, 2024, pp. 5145--5154.

\bibitem{dosovitskiy_carla_2017}
A.~Dosovitskiy, G.~Ros, F.~Codevilla, A.~M. López, and V.~Koltun, ``{CARLA}: {An} {Open} {Urban} {Driving} {Simulator},'' in \emph{1st {Annual} {Conference} on {Robot} {Learning}, {CoRL} 2017, {Mountain} {View}, {California}, {USA}, {November} 13-15, 2017, {Proceedings}}, ser. Proceedings of {Machine} {Learning} {Research}, vol.~78.\hskip 1em plus 0.5em minus 0.4em\relax PMLR, 2017, pp. 1--16.

\bibitem{saleh_voxelscape_2023}
K.~Saleh, M.~Hossny, A.~Abobakr, M.~Attia, and J.~Iskander, ``{VoxelScape}: {Large} {Scale} {Simulated} {3D} {Point} {Cloud} {Dataset} of {Urban} {Traffic} {Environments},'' \emph{IEEE Transactions on Intelligent Transportation Systems}, vol.~24, no.~9, pp. 9435--9448, 2023.

\bibitem{marcus_synth_2025}
R.~Marcus, C.~Vogel, I.~Jatzkowski, N.~Knoop, and M.~Stamminger, ``Synth {It} {Like} {KITTI}: {Synthetic} {Data} {Generation} for {Object} {Detection} in {Driving} {Scenarios},'' 2025.

\bibitem{guillard_learning_2022}
B.~Guillard, S.~Vemprala, J.~K. Gupta, O.~Miksik, V.~Vineet, P.~Fua, and A.~Kapoor, ``Learning to {Simulate} {Realistic} {LiDARs},'' in \emph{2022 {IEEE}/{RSJ} {International} {Conference} on {Intelligent} {Robots} and {Systems} ({IROS})}, 2022, pp. 8173--8180, iSSN: 2153-0866.

\bibitem{zyrianov_learning_2022}
V.~Zyrianov, X.~Zhu, and S.~Wang, ``Learning to {Generate} {Realistic} {LiDAR} {Point} {Clouds},'' in \emph{Computer {Vision} – {ECCV} 2022: 17th {European} {Conference}, {Tel} {Aviv}, {Israel}, {October} 23–27, 2022, {Proceedings}, {Part} {XXIII}}.\hskip 1em plus 0.5em minus 0.4em\relax Berlin, Heidelberg: Springer-Verlag, 2022, pp. 17--35.

\bibitem{ran_towards_2024}
H.~Ran, V.~Guizilini, and Y.~Wang, ``Towards realistic scene generation with lidar diffusion models,'' in \emph{{IEEE/CVF} Conference on Computer Vision and Pattern Recognition, {CVPR} 2024, Seattle, WA, USA, June 16-22, 2024}.\hskip 1em plus 0.5em minus 0.4em\relax {IEEE}, 2024, pp. 14\,738--14\,748.

\bibitem{dai_lidar_2022}
W.~Dai, S.~Chen, Z.~Huang, Y.~Xu, and D.~Kong, ``\BIBforeignlanguage{en}{{LiDAR} {Intensity} {Completion}: {Fully} {Exploiting} the {Message} from {LiDAR} {Sensors}},'' \emph{\BIBforeignlanguage{en}{Sensors}}, vol.~22, no.~19, p. 7533, 2022, number: 19 Publisher: Multidisciplinary Digital Publishing Institute.

\bibitem{viswanath_reflectivity_2024}
K.~Viswanath, P.~Jiang, and S.~Saripalli, ``Reflectivity {Is} {All} {You} {Need}!: {Advancing} {LiDAR} {Semantic} {Segmentation},'' 2024.

\bibitem{anand_advancing_2025}
V.~Anand, B.~Lohani, G.~Pandey, and R.~Mishra, ``Advancing {LiDAR} {Intensity} {Simulation} {Through} {Learning} {With} {Novel} {Physics}-{Based} {Modalities},'' \emph{IEEE Transactions on Intelligent Transportation Systems}, pp. 1--10, 2025, conference Name: IEEE Transactions on Intelligent Transportation Systems.

\bibitem{hu_rangeldm_2024}
Q.~Hu, Z.~Zhang, and W.~Hu, ``{RangeLDM}: {Fast} {Realistic} {LiDAR} {Point} {Cloud} {Generation},'' in \emph{Computer {Vision} – {ECCV} 2024: 18th {European} {Conference}, {Milan}, {Italy}, {September} 29–{October} 4, 2024, {Proceedings}, {Part} {XLIV}}.\hskip 1em plus 0.5em minus 0.4em\relax Berlin, Heidelberg: Springer-Verlag, 2024, pp. 115--135.

\bibitem{bewley_range_2021}
A.~Bewley, P.~Sun, T.~Mensink, D.~Anguelov, and C.~Sminchisescu, ``Range {Conditioned} {Dilated} {Convolutions} for {Scale} {Invariant} {3D} {Object} {Detection},'' in \emph{Proceedings of the 2020 {Conference} on {Robot} {Learning}}, ser. Proceedings of {Machine} {Learning} {Research}, J.~Kober, F.~Ramos, and C.~Tomlin, Eds., vol. 155.\hskip 1em plus 0.5em minus 0.4em\relax PMLR, 2021, pp. 627--641.

\bibitem{mildenhall_nerf_2020}
B.~Mildenhall, P.~P. Srinivasan, M.~Tancik, J.~T. Barron, R.~Ramamoorthi, and R.~Ng, ``{NeRF}: {Representing} {Scenes} as {Neural} {Radiance} {Fields} for {View} {Synthesis},'' 2020.

\bibitem{kerbl_3d_2023}
B.~Kerbl, G.~Kopanas, T.~Leimkuehler, and G.~Drettakis, ``\BIBforeignlanguage{en}{{3D} {Gaussian} {Splatting} for {Real}-{Time} {Radiance} {Field} {Rendering}},'' \emph{\BIBforeignlanguage{en}{ACM Transactions on Graphics}}, vol.~42, no.~4, pp. 1--14, 2023.

\bibitem{chen_survey_2025}
G.~Chen and W.~Wang, ``A {Survey} on {3D} {Gaussian} {Splatting},'' 2024.

\bibitem{manivasagam_lidarsim_2020}
S.~Manivasagam, S.~Wang, K.~Wong, W.~Zeng, M.~Sazanovich, S.~Tan, B.~Yang, W.~Ma, and R.~Urtasun, ``Lidarsim: Realistic lidar simulation by leveraging the real world,'' in \emph{2020 {IEEE/CVF} Conference on Computer Vision and Pattern Recognition, {CVPR} 2020, Seattle, WA, USA, June 13-19, 2020}.\hskip 1em plus 0.5em minus 0.4em\relax {IEEE}, 2020, pp. 11\,164--11\,173.

\bibitem{li_pcgen_2023}
C.~Li, Y.~Ren, and B.~Liu, ``{PCGen}: {Point} {Cloud} {Generator} for {LiDAR} {Simulation},'' in \emph{2023 {IEEE} {International} {Conference} on {Robotics} and {Automation} ({ICRA})}, 2023, pp. 11\,676--11\,682.

\bibitem{huang_neural_2023}
S.~Huang, Z.~Gojcic, Z.~Wang, F.~Williams, Y.~Kasten, S.~Fidler, K.~Schindler, and O.~Litany, ``Neural lidar fields for novel view synthesis,'' in \emph{{IEEE/CVF} International Conference on Computer Vision, {ICCV} 2023, Paris, France, October 1-6, 2023}.\hskip 1em plus 0.5em minus 0.4em\relax {IEEE}, 2023, pp. 18\,190--18\,200.

\bibitem{tao_lidar-nerf_2024}
T.~Tao, L.~Gao, G.~Wang, Y.~Lao, P.~Chen, H.~Zhao, D.~Hao, X.~Liang, M.~Salzmann, and K.~Yu, ``{LiDAR}-{NeRF}: {Novel} {LiDAR} {View} {Synthesis} via {Neural} {Radiance} {Fields},'' in \emph{Proceedings of the 32nd {ACM} {International} {Conference} on {Multimedia}}, ser. {MM} '24.\hskip 1em plus 0.5em minus 0.4em\relax New York, NY, USA: Association for Computing Machinery, 2024, pp. 390--398, event-place: Melbourne VIC, Australia.

\bibitem{zhang_nerf-lidar_2024}
J.~Zhang, F.~Zhang, S.~Kuang, and L.~Zhang, ``Nerf-lidar: Generating realistic lidar point clouds with neural radiance fields,'' in \emph{Thirty-Eighth {AAAI} Conference on Artificial Intelligence, {AAAI} 2024, Thirty-Sixth Conference on Innovative Applications of Artificial Intelligence, {IAAI} 2024, Fourteenth Symposium on Educational Advances in Artificial Intelligence, {EAAI} 2014, February 20-27, 2024, Vancouver, Canada}, M.~J. Wooldridge, J.~G. Dy, and S.~Natarajan, Eds.\hskip 1em plus 0.5em minus 0.4em\relax {AAAI} Press, 2024, pp. 7178--7186.

\bibitem{xue_geonlf_2024}
W.~Xue, Z.~Zheng, F.~Lu, H.~Wei, G.~Chen, and C.~Jiang, ``Geonlf: Geometry guided pose-free neural lidar fields,'' in \emph{Proc. of NeurIPS}, A.~Globersons, L.~Mackey, D.~Belgrave, A.~Fan, U.~Paquet, J.~M. Tomczak, and C.~Zhang, Eds., 2024.

\bibitem{wu_dynamic_2024}
H.~Wu, X.~Zuo, S.~Leutenegger, O.~Litany, K.~Schindler, and S.~Huang, ``Dynamic lidar re-simulation using compositional neural fields,'' in \emph{{IEEE/CVF} Conference on Computer Vision and Pattern Recognition, {CVPR} 2024, Seattle, WA, USA, June 16-22, 2024}.\hskip 1em plus 0.5em minus 0.4em\relax {IEEE}, 2024, pp. 19\,988--19\,998.

\bibitem{tonderski_neurad_2024}
A.~Tonderski, C.~Lindstr{\"{o}}m, G.~Hess, W.~Ljungbergh, L.~Svensson, and C.~Petersson, ``Neurad: Neural rendering for autonomous driving,'' in \emph{{IEEE/CVF} Conference on Computer Vision and Pattern Recognition, {CVPR} 2024, Seattle, WA, USA, June 16-22, 2024}.\hskip 1em plus 0.5em minus 0.4em\relax {IEEE}, 2024, pp. 14\,895--14\,904.

\bibitem{hess_splatad_2025}
G.~Hess, C.~Lindström, M.~Fatemi, C.~Petersson, and L.~Svensson, ``{SplatAD}: {Real}-{Time} {Lidar} and {Camera} {Rendering} with {3D} {Gaussian} {Splatting} for {Autonomous} {Driving},'' 2024.

\bibitem{chen_lidar-gsreal-time_2025}
Q.~Chen, S.~Yang, S.~Du, T.~Tang, P.~Chen, and Y.~Huo, ``{LiDAR}-{GS}:{Real}-time {LiDAR} {Re}-{Simulation} using {Gaussian} {Splatting},'' 2024.

\bibitem{zhou_lidar-rt_2025}
C.~Zhou, L.~Fu, S.~Peng, Y.~Yan, Z.~Zhang, Y.~Chen, J.~Xia, and X.~Zhou, ``{LiDAR}-{RT}: {Gaussian}-based {Ray} {Tracing} for {Dynamic} {LiDAR} {Re}-simulation,'' in \emph{Proceedings of the {IEEE}/{CVF} {Conference} on {Computer} {Vision} and {Pattern} {Recognition}}, 2025.

\bibitem{moenne-loccoz_3d_2024}
N.~Moenne-Loccoz, A.~Mirzaei, O.~Perel, R.~de~Lutio, J.~Martinez~Esturo, G.~State, S.~Fidler, N.~Sharp, and Z.~Gojcic, ``{3D} {Gaussian} {Ray} {Tracing}: {Fast} {Tracing} of {Particle} {Scenes},'' \emph{ACM Trans. Graph.}, vol.~43, no.~6, pp. 232:1--232:19, 2024.

\bibitem{hough_method_1962}
P.~V.~C. Hough, ``\BIBforeignlanguage{English}{{METHOD} {AND} {MEANS} {FOR} {RECOGNIZING} {COMPLEX} {PATTERNS}},'' Originating Research Org. not identified, Tech. Rep. US 3069654, 1962.

\bibitem{duda_use_1972}
R.~O. Duda and P.~E. Hart, ``Use of the {Hough} transformation to detect lines and curves in pictures,'' \emph{Commun. ACM}, vol.~15, no.~1, pp. 11--15, 1972.

\bibitem{ballard_generalizing_1981}
D.~H. Ballard, ``Generalizing the {Hough} transform to detect arbitrary shapes,'' \emph{Pattern Recognition}, vol.~13, no.~2, pp. 111--122, 1981.

\bibitem{biavati_correction_2011}
G.~Biavati, G.~D. Donfrancesco, F.~Cairo, and D.~G. Feist, ``\BIBforeignlanguage{EN}{Correction scheme for close-range lidar returns},'' \emph{\BIBforeignlanguage{EN}{Applied Optics}}, vol.~50, no.~30, pp. 5872--5882, 2011, publisher: Optica Publishing Group.

\bibitem{fang_intensity_2015}
W.~Fang, X.~Huang, F.~Zhang, and D.~Li, ``Intensity {Correction} of {Terrestrial} {Laser} {Scanning} {Data} by {Estimating} {Laser} {Transmission} {Function},'' \emph{IEEE Transactions on Geoscience and Remote Sensing}, vol.~53, no.~2, pp. 942--951, 2015.

\bibitem{schreck_height_2023}
S.~Schreck, H.~Reichert, M.~Hetzel, K.~Doll, and B.~Sick, ``Height {Change} {Feature} {Based} {Free} {Space} {Detection},'' in \emph{2023 11th {International} {Conference} on {Control}, {Mechatronics} and {Automation} ({ICCMA})}, 2023, pp. 171--176, iSSN: 2837-5149.

\bibitem{noauthor_ispc-lablidar4d_2025}
\BIBentryALTinterwordspacing
``ispc-lab/{LiDAR4D},'' Apr. 2025, original-date: 2024-03-06T07:39:59Z. [Online]. Available: \url{https://github.com/ispc-lab/LiDAR4D}
\BIBentrySTDinterwordspacing

\bibitem{zhang_unreasonable_2018}
R.~Zhang, P.~Isola, A.~A. Efros, E.~Shechtman, and O.~Wang, ``The unreasonable effectiveness of deep features as a perceptual metric,'' in \emph{2018 {IEEE} Conference on Computer Vision and Pattern Recognition, {CVPR} 2018, Salt Lake City, UT, USA, June 18-22, 2018}.\hskip 1em plus 0.5em minus 0.4em\relax {IEEE} Computer Society, 2018, pp. 586--595.

\bibitem{wang_image_2004}
Z.~Wang, A.~Bovik, H.~Sheikh, and E.~Simoncelli, ``Image quality assessment: from error visibility to structural similarity,'' \emph{IEEE Transactions on Image Processing}, vol.~13, no.~4, pp. 600--612, 2004.

\bibitem{fan_point_2017}
H.~Fan, H.~Su, and L.~J. Guibas, ``A point set generation network for 3d object reconstruction from a single image,'' in \emph{2017 {IEEE} Conference on Computer Vision and Pattern Recognition, {CVPR} 2017, Honolulu, HI, USA, July 21-26, 2017}.\hskip 1em plus 0.5em minus 0.4em\relax {IEEE} Computer Society, 2017, pp. 2463--2471.

\end{thebibliography}

\end{document}